\documentclass[sigplan,10pt]{acmart}
\renewcommand\footnotetextcopyrightpermission[1]{}
\usepackage{booktabs}
\usepackage{subfigure}
\usepackage{makecell}
\usepackage{algorithm}
\usepackage{algorithmic}
\usepackage{amsmath}
\usepackage{array} 
\usepackage{adjustbox}
\usepackage{multirow}
\usepackage{newfloat}
\usepackage{listings}
\usepackage{pifont}
\AtBeginDocument{%
	}

\newcommand{\pname}[1]{{Adacc}{#1}}

\begin{document}
	
\title{\pname{}: An Adaptive Framework Unifying Compression and Activation Recomputation for LLM Training}
	\newcommand{\cp}[1]{{\color{black} #1}}
	\author{{Ping Chen$^{125}$*, Zhuohong Deng$^{3}$*, Ping Li$^{4}$, Shuibing He$^{125}$, Hongzi Zhu$^{3}$, Yi Zheng$^{4}$, Zhefeng Wang$^{4}$, Baoxing Huai$^{4}$, Minyi Guo$^{3}$}\\
		{$^1$The State Key Laboratory of Blockchain and Data Security, Zhejiang University\\
			$^2$Hangzhou High-Tech Zone (Binjiang) Institute of Blockchain and Data Security,\\ $^3$Shanghai Jiao Tong University,  $^4$Huawei Cloud, $^5$Zhejiang Lab}\\
		$^{*}$Equal contribution \\
}

	\begin{abstract}

	Training large language models (LLMs) is often constrained by GPU memory limitations. To alleviate memory pressure, activation recomputation and data compression have been proposed as two major strategies. However, both approaches have limitations: recomputation introduces significant training overhead, while compression can lead to accuracy degradation and computational inefficiency when applied naively. 

In this paper, we propose \pname{}, the first adaptive memory optimization framework that unifies activation recomputation and data compression to improve training efficiency for LLMs while preserving model accuracy. Unlike existing methods that apply static, rule-based strategies or rely solely on one technique, Adacc makes fine-grained, tensor-level decisions, dynamically selecting between recomputation, retention, and compression based on tensor characteristics and runtime hardware constraints.
\pname{} tackles three key challenges: (1) it introduces layer-specific compression algorithms that mitigate accuracy loss by accounting for outliers in LLM activations; (2) it employs a MILP-based scheduling policy to globally optimize memory strategies across layers; and (3) it integrates an adaptive policy evolution mechanism to update strategies during training in response to changing data distributions. Experimental results show that Adacc improves training throughput by 1.01$\times$ to 1.37$\times$ compared to state-of-the-art frameworks, while maintaining accuracy comparable to the baseline.

	\end{abstract}

	\keywords{Large Model Training, Memory Optimization}

	\settopmatter{printfolios=true}
	\maketitle
	\footnotetext[1]{Work done during the internships of  Ping Chen and Zhuohong Deng at Huawei Cloud, both of whom contributed equally to this work.}
	
	\pagestyle{plain}

\section{Introduction}
\label{sec:intro}

Large Language Models (LLM) have achieved unprecedented success across various domains. Scaling laws show that model size is crucial for performance, leading to increasingly larger models. For instance, the size of large models has increased by over 360×, from GPT-2~\cite{gpt2} with 1.5 billion parameters to PaLM with 540 billion parameters~\cite{palm}. This trend is expected to continue, with model sizes exceeding the memory capacity of a single GPU (tens of gigabytes).

To address GPU memory limitations for training large models, the recomputation approach (activation checkpointing) has been proposed. It alleviates memory pressure by discarding activations during forward propagation and regenerating them on demand during backward propagation~\cite{checkpointing-arxiv16}. This mainstream method is widely used in frameworks like Megatron~\cite{megatron} and MindSpore~\cite{MindSpore}, which implement various policies to manage which tensors are retained or recomputed.

Existing recomputation methods often incur substantial training overhead due to their rigid, rule-based design~\cite{lynx}. These approaches typically select specific tensors for recomputation based on predefined heuristics, following an all-or-nothing pattern~\cite{Deepspeed-Checkpointing}. For example, Megatron-LM provides a \textit{full recomputation} mode, which retains only the inputs to transformer layers as checkpoints and discards all other activations, requiring them to be recomputed during the backward pass.
Specifically, we profile Megatron-LM using a 345M parameter GPT model with an 8 batch size on 4 Tesla V100 GPUs. Our results indicate that full recomputation accounts for over 28\% of the total training time. This inefficiency arises because the rule-based policy does not account for the varying memory demands and computational costs of different model layers.


\begin{table}
	\centering 
	\caption{The comparisons of different techniques.} 
	\footnotesize
	\begin{center}
		\begin{tabular}{m{0.6cm}<{\centering}m{0.5cm}<{\centering}m{1.1cm}<{\centering}m{1.2cm}<{\centering}m{1.5cm}<{\centering}m{0.8cm}<{\centering}} \hline
			\textit{Tensor}        & \textit{Size (MB)}  & \textit{Recmp. time (ms)} & \textit{Compression time (ms)}& \textit{Size after compression (MB)} & \textit{Recmp.\textcolor{blue}{\ensuremath{\clubsuit}} Comp.\textcolor{red}{\ensuremath{\bigstar}} } \\ \hline
			T1   &  96               &  \textbf{0.36}              & 0.37 & 24 &\textcolor{blue}{\ensuremath{\clubsuit}}  \\ \hline
			T2	  &  42  &	1.02	            & \textbf{0.16} & 11.8 &\textcolor{red}{\ensuremath{\bigstar}} \\ \hline
			T3 &  42     &  0.58      & \textbf{0.16} & 11.8&\textcolor{red}{\ensuremath{\bigstar}} \\ \hline
			T4  &  10.5               &  \textbf{0.04}           & 0.04 & 2.6&\textcolor{blue}{\ensuremath{\clubsuit}}  \\ \hline
		\end{tabular}
		
		*Recmp. time: the time required to recompute from the previous layer.\\
		\label{tbl:compression}
		\vspace{-0.2in}
	\end{center}
\end{table}

In addition to recomputation, data compression has emerged as another mainstream approach for reducing memory usage during model training. A common strategy is to apply quantization to compress activations after the forward pass and decompress them before the backward pass, thereby reducing memory consumption without modifying the computational graph~\cite{chen2021actnn}.
However, even simple quantization schemes introduce non-negligible overhead due to the additional computation required for both quantization and dequantization, which can negatively impact overall training efficiency.

We find that different activation tensors are best optimized using different memory strategies—some favor recomputation, while others benefit more from compression. This variation arises from intrinsic differences in tensor characteristics, such as their size and computational cost. To validate this, we select several representative tensors from a GPT model and compare their memory and time trade-offs under both optimization methods.

As shown in Table~\ref{tbl:compression}, we evaluate four representative tensors, each favoring a different memory optimization strategy. 
To quantify the trade-offs, we define the optimization bandwidths, $Band_{recomp}$ and $Band_{comp}$, as the \textit{saved memory per unit time}, calculated as
$\frac{\text{Tensor}_{\text{size}} - \text{Storage}_{\text{used}}}{\text{Time}}$,
where $\text{Storage}_{\text{used}}$ denotes the actual memory footprint under each strategy (e.g., zero for recomputation and compressed size for compression). This metric reflects the memory-efficiency of each technique in terms of throughput.
For instance, T1 yields a $Band_{recomp}$ of 266 and a $Band_{comp}$ of 194, indicating that recomputation is more favorable. In contrast, T2 exhibits higher efficiency under compression, making it the better choice for that tensor.

These results demonstrate that applying a one-size-fits-all policy—whether solely relying on recomputation or compression—is suboptimal. Instead, \textit{tensor-aware hybrid strategies} that dynamically adapt to the properties of individual tensors are essential for maximizing training throughput, challenging the static, rule-based approaches employed in mainstream frameworks such as Megatron~\cite{megatron}.


This insight motivates an adaptive framework combining compression and recomputation to maximize training efficiency and mitigate GPU memory limitations.
However, it faces three key challenges. First, directly applying compression may lead to significant accuracy degradation. Second, it is non-trivial to determine the optimal combination of compression and recomputation strategies within a reasonable time budget. Third, the effectiveness of compression depends on the statistical properties of activations, such as the presence of outliers, which can vary dynamically during training. As a result, a fixed compression policy may become suboptimal as training progresses.

To address these challenges, we propose \pname{}, which first introduces four layer-specific compression algorithms tailored to different types of activations; notably, one of them explicitly handles data outliers to improve compression robustness, while others adopt lightweight strategies for efficiency. It then employs a custom mixed-integer linear programming (MILP) algorithm to derive globally optimized tensor-level policies, significantly reducing the search space by leveraging the repetitive structure of LLMs. Finally, \pname{} adaptively updates its optimization policy during training to accommodate dynamic data characteristics and sustain high throughput.

We conduct extensive evaluations of \pname{} on a broad range of benchmarks and downstream tasks, showing that it consistently improves training performance without compromising model accuracy.



\section{ Background and Motivation}
\label{sec:bg}

\subsection{Large language model training}
Large language models, composed of multiple repeated layers  (e.g., Transformer block), are trained iteratively with data batches. Each step involves forward propagation (FP) to compute large activations and backward propagation (BP) to calculate gradients for optimization. 
A model with M parameters requires 16M + $\delta$M bytes of memory, including 2M FP16 parameters, 2M  FP16 gradients, 12M  FP32 optimizer data (e.g., Adam), and $\delta$M FP16 intermediate activation, where $\delta$ depends on the user-defined batch size.
Generally,
training samples are typically processed in large batches to enhance throughput and device utilization by increasing arithmetic density~\cite{lynx}, leading to substantial GPU memory usage for activations.

\begin{figure}
	\centering
	\includegraphics[width=3.2in]{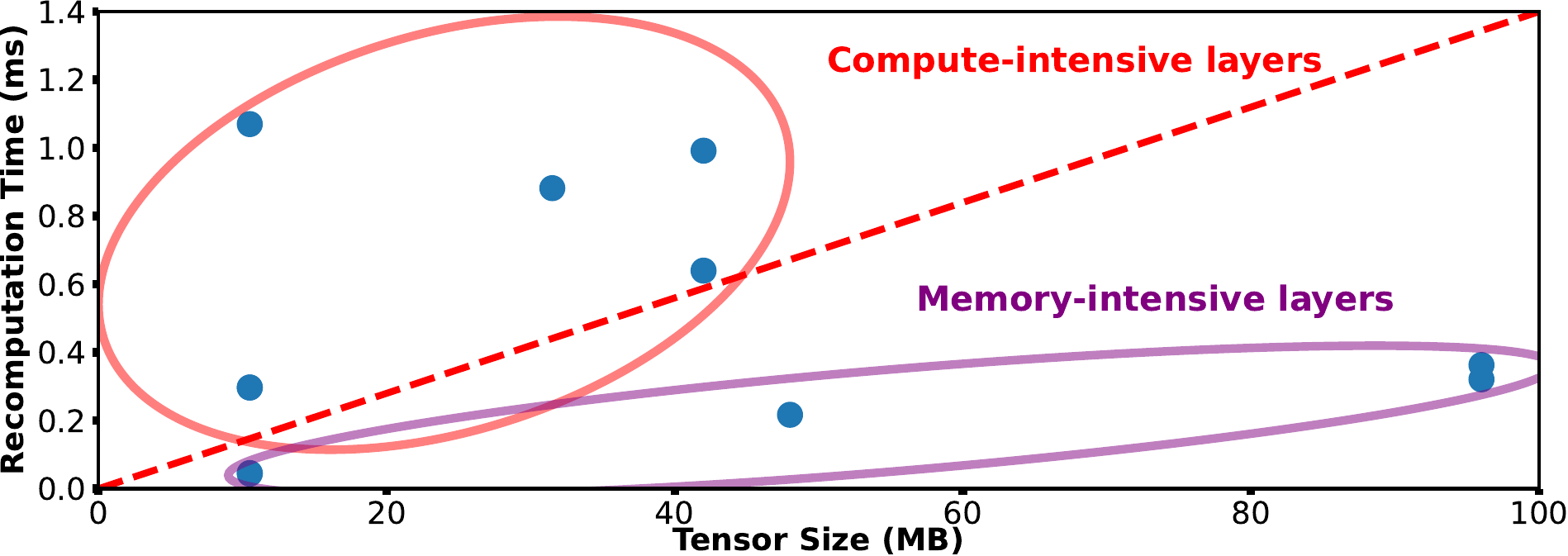}
	\caption{The analysis of layers (tensors) in a Transformer block of GPT model. The red circle indicates the compute-intensive layer, while the purple represents the memory-intensive layer, better suited for recomputation.}
	\label{fig:moti}
\end{figure}

However, GPUs' limited memory capacity restricts large model training, as both model states and activations (feature maps) consume memory~\cite{Zero-SC20}. Model states consist of parameters, gradients, and optimizer states, such as momentum and variance in algorithms like Adam~\cite{Adam}.
Activations are intermediate data generated during training that need to be stored on the GPU, with memory consumption increasing as batch size grows.
Users often use large training batches to maximize GPU utilization,  which exacerbates memory bottleneck issue (i.e., memory wall).
For example, in GPT model training, 66\% of memory is consumed by stashed immediately activations on 4 V100 GPUs with a batch size of 4. This number grows to 76\% for 8 batch size.
Therefore, optimizing activations is essential for efficient LLM training.

\subsection{Existing works and their shortcomings}
\label{sec:mot}
\textbf{Recomputation.} Several studies have addressed the GPU memory wall issue by exploring various optimization techniques aimed at alleviating the constraints imposed by limited GPU memory. One common approach involves activation checkpointing (recomputation), which temporarily discards intermediate activations and re-computes them during the backward pass to save memory~\cite{checkpointing-arxiv16, checkmate19, lynx}. While this method is effective in extending GPU memory capacity, we observe that not all activations are suitable for recomputation. Specifically, small activations that are costly to recompute should either remain in GPU memory or be optimized using alternative techniques, as highlighted by the tensors within the red circle in Figure~\ref{fig:moti}. These activations present unique challenges that are often overlooked in traditional checkpointing strategies.

\textbf{Compression.} Further attempts to address memory limitations have involved both lossless and lossy encoding techniques, which compress activation data to reduce memory consumption~\cite{GIST-ISCA18, cDMA-HPCA18}. However, these approaches are typically designed for traditional computer vision (CV) models, and they do not directly cater to the unique memory demands of large language models (LLMs), which have more complex and larger activation patterns. In contrast, \pname{} is specifically tailored to optimize memory usage in the context of LLMs, addressing these specific challenges with greater efficiency.

\textbf{Data swapping.} Another approach involves swapping activations between GPU and CPU memory, which can provide additional memory space on the GPU~\cite{VDNN-MICRO16, CSWAP+-TPDS22}. However, this technique introduces significant performance overhead due to the limited PCIe bandwidth, and often requires intrusive modifications to the underlying framework, making it impractical for real-time or large-scale training scenarios~\cite{liao2024lohan}.

\textbf{Parallelism policy.}In addition, methods that leverage intra- and inter-operator parallelism have been proposed to distribute model partitions across multiple GPUs, thereby reducing the memory footprint on each individual GPU~\cite{Zero-SC20, Megatron-Arxiv19, Gpipe-NIPS19}. These methods complement \pname{}, providing an alternative approach to memory optimization by spreading the computational load. However, they still face challenges in terms of coordination and communication overhead between GPUs.

In summary, while various methods have been proposed to address the GPU memory wall issue, each comes with trade-offs in terms of computational cost, practicality, and the specific requirements of large-scale models. \pname{} distinguishes itself by offering a more tailored solution for the memory optimization needs of large language models.


\section{Design}
\label{sec:design}

\begin{figure}
	\centering
	\includegraphics[width=3.5in]{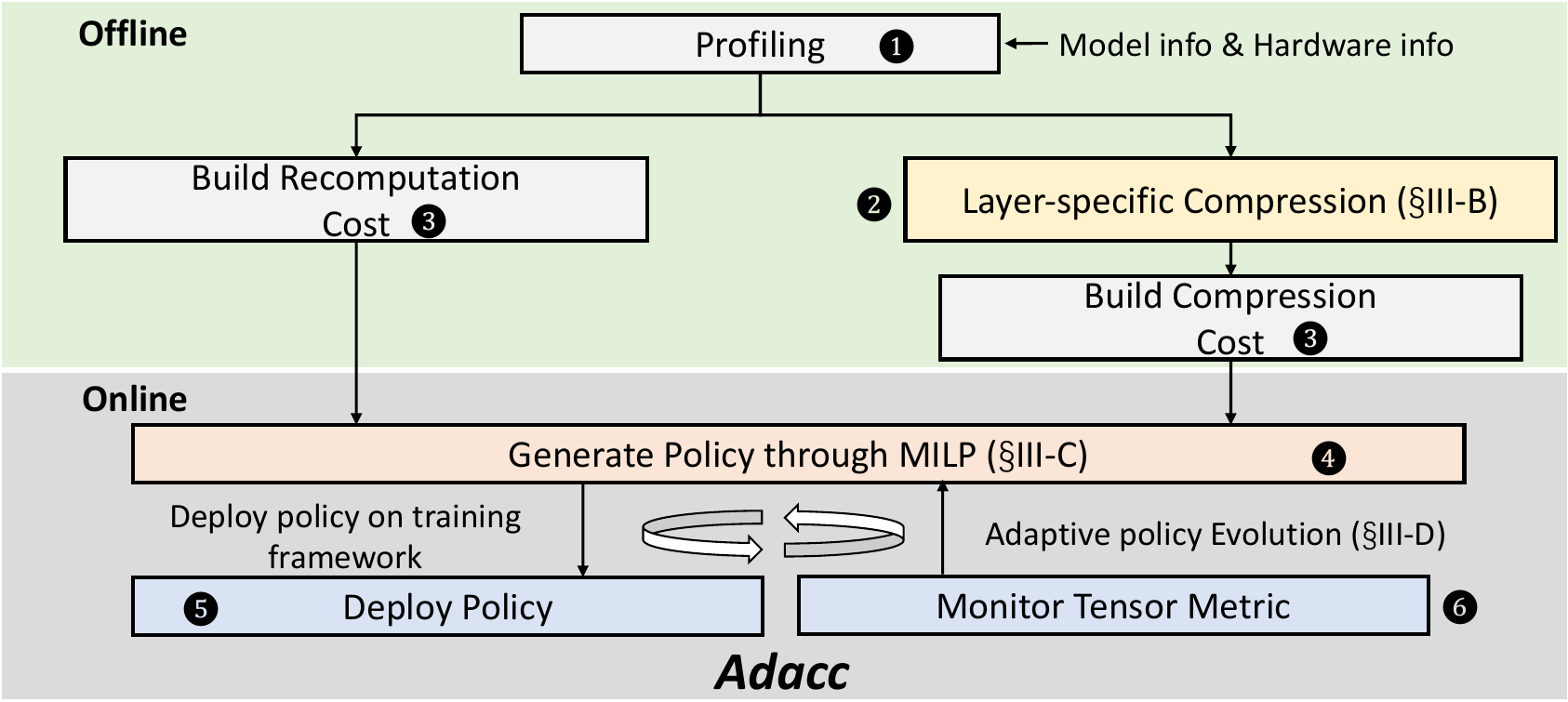}
	\caption{The overview of \pname{}. 
	}
	\label{fig:overview}
\end{figure}

\pname{} aims to train LLMs on one or more GPUs,  when GPU memory or a user-specified budget is insufficient. It keeps the training parameters, such as batch size, unchanged while optimizing performance and preserving accuracy.
At the high level, \pname{} adopts the idea of adaptive activation checkpointing and compression. 
While, it introduces three challenges.
(1) How to design compression algorithms that minimize accuracy loss? (2) How to optimize memory usage with compression and recomputation? (3) How to dynamically adjust policies during training to maximize throughput?

\subsection{Overview}
\label{sec:overview}

\begin{figure*}
	\centering
	\includegraphics[width=7in]{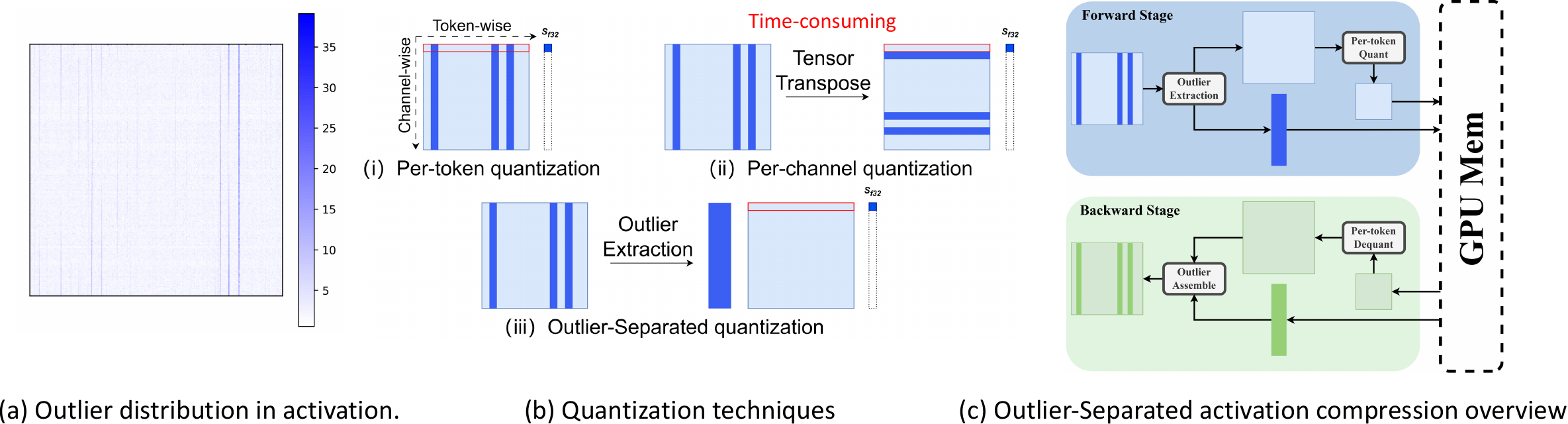}
	\caption{(a) Outlier distribution (dark blue) of a tensor in GPT. We observe that the outliers are grouped along the channel dimension.
	(b) The illustration of different compression techniques. $S_{f32}$ represents the scaling factor for each row of data.
	(c) The  workflow of outlier-separated compression.}
	\label{fig:compression}
\end{figure*}

\begin{table}[t]
	\centering
	\caption{Variables used in MILP algorithm.}
	\begin{tabular}{p{2cm}p{5.5cm}}  
		\toprule
		\multicolumn{2}{l}{\textbf{Constant variables:}} \\
		$N$ & Number of operators \\
		$M_{constraint}$ & GPU memory capacity \\
		$M_{static}$ & Fixed memory used for storing static data (e.g., model states, gradients, and optimizers) \\
		$N_{Layer}$ & Number of transformer layers \\
		
		\multicolumn{2}{l}{\textbf{Optimization variables:}} \\
		$R_{i,t}$ & Boolean variable. If $R_{i,t}$ is true, it means $op_i$ will follow this policy \\
		
		\multicolumn{2}{l}{\textbf{Intermediate variables:}} \\
		$M_i$ & Memory size for $op_i$ \\
		$Tcomp_i$ & Computation time for $op_i$ \\
		$Tc_i$ & Compression time for $op_i$ \\
		$Tdc_i$ & Decompression time for $op_i$ \\
		$CRate_i$ & Compression Rate for $op_i$ \\
		\bottomrule
	\end{tabular}
	\label{tbl:milp}
\end{table}

\pname{} consists of three key modules: the model profiler, policy maker, and model rectifier, which collaborate to achieve the design goals. Figure~\ref{fig:overview} provides an overview of \pname{}.

\textbf{Model profiler.} Before deploying a model, a test run is conducted with user-defined configurations (GPU features, model architecture, batch size, etc.). Metrics like execution time, operator size, and dependencies are recorded in a database for the policy maker's scheduling decisions \ding{182}.
Based on the profiled tensor size and our defined compression algorithm, we calculate the compression rate and the time cost for both compression and decompression \ding{183}.

\textbf{Policy maker.} 
Next, we input the collected operator size, computation time, tensor compression rate, and compression/decompression time into the policy maker's MILP cost model \ding{184} and determine the optimization policy for each tensor \ding{185}.
The training framework then implements the optimal schedule defined by the model policy maker, deploying the model for training on physical devices using the training framework \ding{186}.

\textbf{Policy rectifier.} 
The monitor periodically collects data characteristics during training to detect if the policy is suboptimal and adjusts it promptly \ding{187}.

\subsection{Layer-specific Compression}
\label{sec:compression}

\textbf{Traditional compressions.}
A straightforward approach is to compress activations from FP16 to INT4 using per-token quantizaiton (as shown in Figure~\ref{fig:compression}(b)(i)).
However, 
our experience shows that this naive approach can reduce model accuracy by up to 39\% in GPT models or even leads to gradient overflow in training (as shown in evaluation section).
To solve this challenge, we design several specific compression algorithm for different LLM layers.

\textbf{Outlier-separated activation compression.}
As illustrated in Figure~\ref{fig:compression}(a), activations in LLMs often exhibit a heavy-tailed distribution, with a small number of extreme outlier values. These outliers can carry important semantic or gradient information, particularly in attention and normalization layers. However, existing quantization methods typically apply uniform quantization across the entire tensor without isolating outliers, forcing them to share quantization levels with densely populated regions of the value distribution. This results in high quantization error for the outliers, leading to loss of critical information and significant model degradation. 

To address this, a \textit{naive method} groups activations by channel and quantizes outliers separately from the rest of the values, as shown in Figure~\ref{fig:compression}(b)(ii). However, since the data is primitively stored along the token dimension (row), data adjacent on the channel dimension (column) is not contiguous in memory. Since the compressor requires contiguous data in memory, we need to transpose and align the tensor before compression. Unfortunately, performing these operations can be extremely resource-intensive and significantly slow down the process.
Through our observations, we noticed that isolating outliers from normal values significantly improves efficiency. This is because, by separating the outliers, we can avoid the need for tensor transposition and alignment, which are computationally expensive. Without the need to rearrange the data in memory, the normal values can be quantized directly while maintaining their original alignment, thus reducing the overhead.

Based on this insight, we propose an \textit{outlier-separated method} that isolates the outliers from the normal values and quantizes the normal values without the need for additional transformations, as shown in Figure~\ref{fig:compression}(b)(iii). This approach streamlines the process and avoids the performance bottleneck caused by tensor manipulation.

To efficiently isolate outlier channels from activations, we designed a Z-Score method. The Z-Score is a statistical measure that quantifies how far a particular data point (or in this case, an activation channel) deviates from the mean in terms of standard deviations. This method is crucial for distinguishing extreme values from the normal distribution of the data.
The Z-Score for each channel is calculated using the following formula:

\begin{equation}
	\label{eq:z-score}
	Z=\frac{S_{h_{i}-\mu}}{\sigma}
\end{equation}
where $S_{h_{i}}$ is the absolute summation of $i$th channel, $\mu$ and $\sigma$ represent the average absolute summations,
 and the standard deviation of absolute summations respectively. We set a Z-Score threshold to extract outlier channels,
and it is 3 in our work (experiments show that this setting achieves the best model accuracy). Outlier channels are set to zero in original activation.
The extracted outlier channels and their indices are stored without compression to maintain model accuracy, while the normal activations are compressed from FP16 to INT4 to reduce memory usage. The compression and decompression workflows are shown as Figure~\ref{fig:compression}(c).

\textbf{Activation quantization.}
We regard tensors of arbitrary dimensions as one-dimensional array stored continuously in memory, and group the activations.
  For \textbf{symmetric quantization}, in each quantization group,
   scaling factor is calculated by $S_{f32}=\frac{\max(|X_{f16}|)}{8}$,
    where $|X_{f16}|$ is the max absolute value,
	 and $8$ is the max absolute value 4bit signed integer can represented. After that, each value is calculated by formulation:

\begin{equation}
	\label{eq:symmetric-quant}
	X_{i4}=clip(\left\lfloor(\frac{X_{f16}}{S_{f32}})\right\rceil, -8, 7)
\end{equation}

where $\left\lfloor\right\rceil$ means nearest rounding, $X_{i4}$ is the quantized value, and \textit{clip} ensures the $X_{i4}$ greater than 7 are capped at 7. During dequantization, the activations are restored using formulation:
$\hat{X}_{f16}=X_{i4}\cdot S_{f32}$. \textit{For majority of activations, which are distributed on both side of zero, symmetric quantizaiton can be applied to accelerate quantizaiton (e.g., T=[-2,-1,0,1,2]).}

For \textbf{asymmetric quantization}, we calculate offset for each group by $O_{f32}=\frac{\max(X_{f16})+\min(X_{f16})}{2}$,
and the scaling factor is calculated using 
$S_{f32}=\frac{\max(X_{f16})-\min(X_{f16})}{16}$,
where $\max(X_{f16})$ and $\min(X_{f16})$ is the maximum and the minmum value in the group, respectively. Each value is calculated by formulation:

\begin{equation}
	\label{eq:asymmetric-quant}
	X_{i4}=clip(\left\lfloor(\frac{X_{f16}-O_{f32}}{S_{f32}})\right\rceil, -8, 7)
\end{equation}
during dequantization, the activation are restored using formulation:
$\hat{X}_{f16}=X_{i4}\cdot S_{f32}+O_{f32}$. \textit{Asymmetric quantizaiton are applied to the activations, which distributes on only one side of zero to minimize quantizaiton errors (e.g., T=[1,2,3]).}

\textbf{Layer-specific compression.}
We apply different compression schemes to activations according to their statistical and computational properties. Importantly, the targeted layer types are core components in most large language models, including GPT, BERT, T5, and LLaMA, which makes our method architecture-agnostic and widely applicable.

\begin{itemize}

\item For \textit{Linear}, \textit{LayerNorm}, and \textit{ } layers with many outliers, we apply an outlier-separated activation compression scheme to minimize errors.

	
\item For the \textit{Query, Key, and Value matrixs}, we using channel-wise symmetric quantization, as no extra memory alignment is needed.

\item For the \textit{Softmax} layer and \textit{Score}, with activations distributed between (0, 1) and no outliers, we apply asymmetric quantization to save ensure accuracy.

\item For the \textit{Dropout mask} in attention layers, we compress the original byte format to bit format without accuracy loss.

\end{itemize}




\subsection{Determining Cost-Effectiveness of Tensor Optimization}
\label{sec:policy}

\begin{figure}
	\centering
	\includegraphics[width=3.2in]{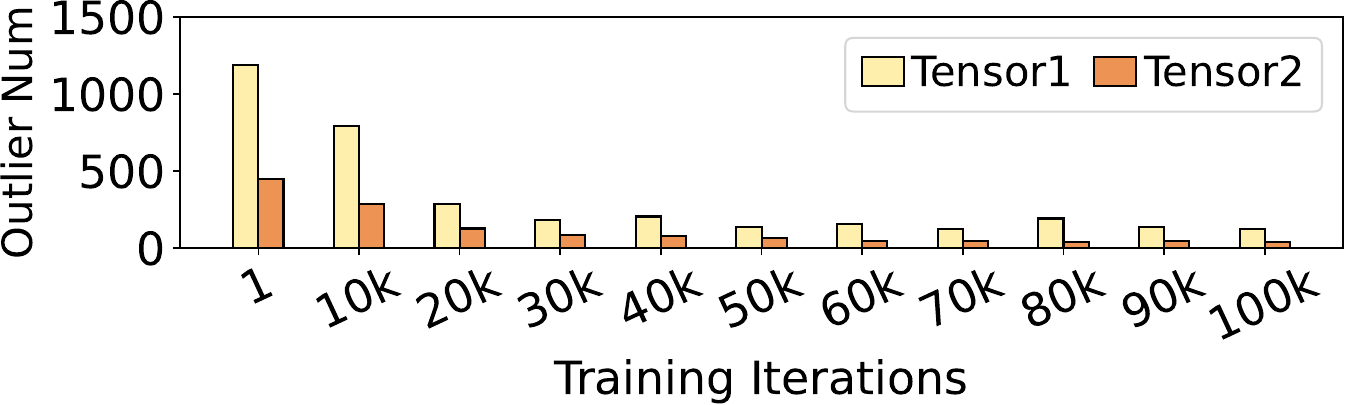}
	\caption{The number of outliers fluctuates during the training of two tensors in GPT. Fewer outliers lead to a higher data compression rate.}
	\label{fig:outlier}
\end{figure}

As analyzed in the motivation section, selecting an appropriate memory optimization policy for different tensors is crucial for achieving optimal training efficiency. Each layer can opt for either layer-specific compression or recomputation. To determine the optimal policy, existing approaches often construct cost models over the full computation graph~\cite{checkmate19}, resulting in a large combinatorial search space that is impractical for large language models (LLMs).

\textbf{Challenges.} Designing this optimization problem presents several challenges. First, recomputation decisions must respect data dependencies: a layer can only be recomputed if all its upstream layers are either stored or also marked for recomputation, introducing nontrivial inter-layer constraints. Second, compression introduces heterogeneous trade-offs, different layers employ distinct compression schemes with varying memory savings and computational overheads, which must be precisely modeled to ensure accurate cost estimation. Together, these factors not only complicate the problem formulation but also significantly enlarge the solution space, making the optimization process more challenging.

Fortunately, LLMs consist of repeated, structurally identical blocks—such as transformer layers in GPT, with consistent memory and data patterns. This regularity allows us to derive a locally optimal policy for one block and apply it across others, significantly reducing the search space without compromising policy quality.

\begin{figure}[t]
	\centering
	\subfigure[GPT-117M.]{
		\includegraphics[width=3in]{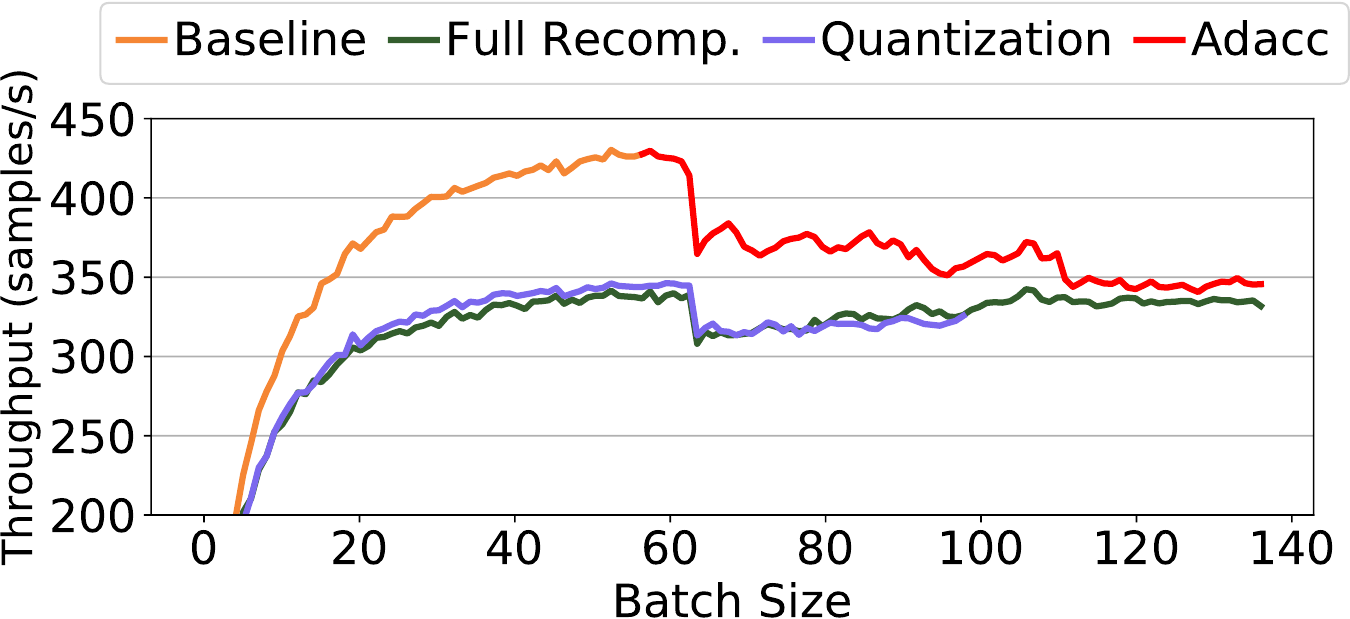}
		\label{fig:100m}
	}
	\subfigure[GPT-345M.]{
		\includegraphics[width=3in]{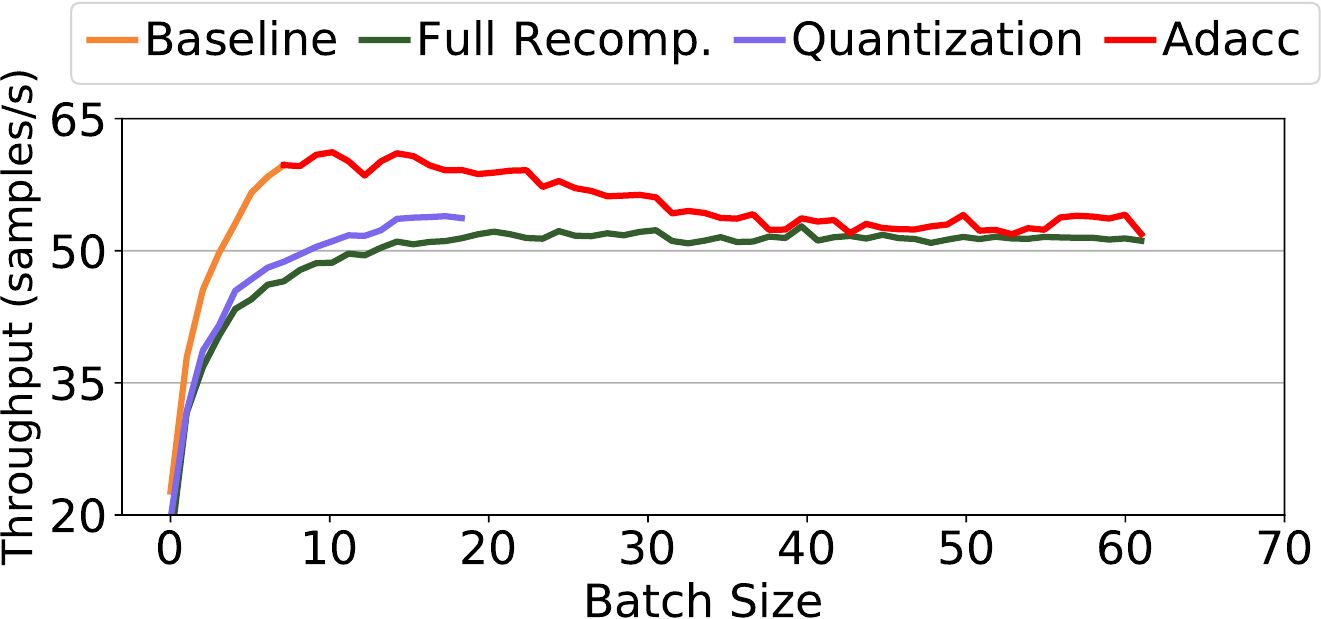}
		\label{fig:300m}  }
	\subfigure[GPT-4.7B.]{
		\includegraphics[width=3in]{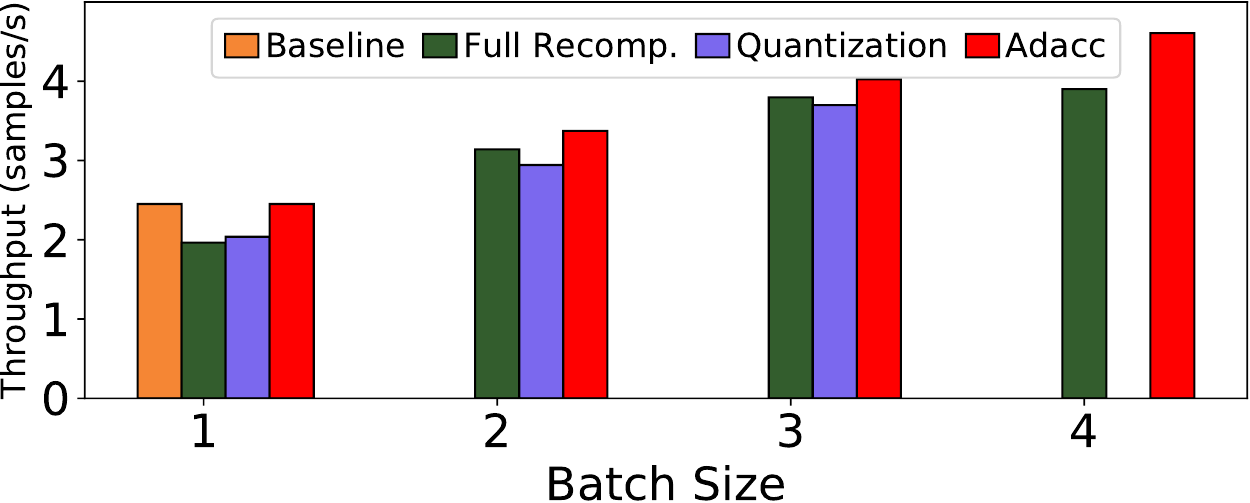}
		\label{fig:throughput_large}  }
	\caption{Overall training throughput for various batch sizes across four different memory optimization techniques. We omit displaying evaluation results that encounter out-of-memory issues.}
	\label{fig:throughput}
\end{figure}

\textbf{MILP cost model.} We formulate the optimization as a mixed-integer linear program (MILP), leveraging the repeated structural patterns commonly found in LLMs to reduce the problem complexity. For illustration, we describe the design using a transformer-based model; however, the approach generalizes to other large-scale deep learning models with similar modular architectures.

\textbf{Problem definition.} The DNN model comprises $N$ operators ($op_1..op_n$) that perform training operation saccording to the model topology.
$R_{i,t}$ represents the optimization policy for $op_i$. $R_{i,0}=1$, $R_{i,1}=1$, and $R_{i,2}=1$ represent recomputation, compression, and retention in GPUs, respectively.
Table~\ref{tbl:milp} summarizes all variables used. 
The objective is to minimize training time (i.e., minimizing the recomputation, compression and decompression overhead) while avoiding out-of-memory issues, which can be formulated as:

\begin{equation}
	\label{eq:objective}
	{\small
		\begin{aligned} 
				& \underset{}{\text{minimize}} & & \sum_{t=1}^{n} Tcomp_i \times R_{i,0} + \sum_{t=1}^{n} (Tc_i + Tdc_i) \times R_{i,1} \\
				& \text{subject to} && \text{Dependency constraints}\\
				& && \text{Memory constraints\:}
		\end{aligned}
	}
\end{equation}

\textbf{Dependency constraints.}
We constrain each optimization operation for every operator to be executed only once in Equation~\ref{eq:cons1}.
Since all recomputations require an initial layer as a checkpoint, we store the output of the first layer of each Transformer as the checkpoint.
Thus, the output of $op_1$ 
is either retained or compressed in the GPU, as shown in  Equation~\ref{eq:cons2}.

\begin{align}
	& \sum_{t=1}^{3} R_{i,t}=1 \quad \forall i \label{eq:cons1} \\
	& R_{1,0}  = 0 \quad \forall t \label{eq:cons2} \\
	& M_{static} + M_{act} \leq M_{constraint}  \label{eq:cons3} \\
	& M_{act} = N_{Layer}  \times  \sum_{t=1}^{n} M_i \times (R_{i,1} \times CRate_i +  R_{i,2}) \label{eq:cons4}  
\end{align}

\textbf{Memory constraints.}
The data stored in the GPU should not exceed the GPU’s memory. GPU memory mainly consists primarily of static memory and activations, so it's crucial to ensure them stays within the GPU's capacity, as shown in Equation~\ref{eq:cons3}.
Activation consists of tensors residing GPU memory ($R_{i,2}=1$) and remaining tensors after compression ($R_{i,1}=1$), as shown in Equation~\ref{eq:cons4}.

\textbf{Solving MILP.}
We generate the recomputation policy by using PulP~\cite{pulp}.
Our design drastically reduces the search space compared to existing approaches that model all operators, taking less than 0.5 second to find an optimal policy, as a negligible time relative to the overall training time.

\begin{table}[t]
	\centering
	\small
	\caption{Maximum batch size.
		\label{tbl:maxbatch}}
	\setlength{\tabcolsep}{1mm}{
		\resizebox{\linewidth}{!}{
			\begin{tabular}{cccccc}
				\toprule
				\textbf{Model} & \makecell{\textbf{Baseline}} & \makecell{\textbf{Full Recomp.}} & \makecell{\textbf{Quantization}}& \makecell{\textbf{\pname{}} }\\
				\midrule
				GPT-117M 			& 57 & 136 &98 &\textbf{136} \\
				GPT-345M  		& 8 & 61 &19 &\textbf{61} \\
				GPT-4.7B  		& 1 & 4 &3 &\textbf{4} \\
				\bottomrule
			\end{tabular}
		}
	}
\end{table}

\subsection{Adaptive Policy Evolution}
\label{sec:evolution}

\begin{table}[t]
	\caption{Evaluations on downstream tasks.
		\label{tbl:downstream}}
	\begin{adjustbox}{width=\linewidth, center}
		\begin{tabular}{clllllll}
			\hline
			\multicolumn{1}{l}{MODEL}   & BS          & METHOD             & Lambada & Arc-e & Sciq  & Swag  & Ave. \\ \hline
			\multirow{8}{*}{117M} & \multirow{4}{*}{32} & Baseline           & 53.46   & 39.56 & 71.80 & 37.46 & 50.57   \\ \cline{3-8} 
			&                     & Quantization       & 2.77    & 25.08 & -     & 24.66 & 13.13   \\
			&                     & Adacc              & \textbf{53.13}   & 39.1  & \textbf{70.80} & \textbf{37.22} & \textbf{50.06}   \\
			&                     & Adacc  w/o outlier & 53.03   & \textbf{39.18} & 69.90 & 36.98 & 49.77   \\ \cline{2-8} 
			& \multirow{4}{*}{64} & Baseline           & 54.76   & 42.05 & 72.60 & 38.19 & 51.90   \\ \cline{3-8} 
			&                     & Quantization       & 3.14    & 25.08 & -     & 24.66 & 13.22   \\
			&                     & Adacc              & \textbf{54.82}   & \textbf{41.04} & 71.80 & \textbf{38.09} & \textbf{51.44}   \\
			&                     & Adacc  w/o outlier & 54.7    & 39.77 & 71.80  & 38.09 & 51.03   \\ \hline
			\multirow{8}{*}{345M} & \multirow{4}{*}{8}  & Baseline           & 56.97   & 42.85 & 72.50  & 39.49 & 52.95   \\ \cline{3-8} 
			&                     & Quantization       & 43.1    & 25.08 & -     & 24.66 & 23.21   \\
			&                     & Adacc              & \textbf{56.29}   & \textbf{42.55} & \textbf{72.70}  & \textbf{39.26} & \textbf{52.7}    \\
			&                     & Adacc w/o outlier & 56.06   & 41.79 & 71.60  & 39.12 & 52.14   \\ \cline{2-8} 
			& \multirow{4}{*}{16} & Baseline           & 58.06   & 44.65 & 76.20  & 40.88 & 54.95   \\ \cline{3-8} 
			&                     & Quantization       & 50.45   & 25.08 & -     & 24.65 & 25.05   \\
			&                     & Adacc              & \textbf{57.67}   & \textbf{44.53} & \textbf{76.60}  & \textbf{40.79} & \textbf{54.90}   \\
			&                     & Adacc  w/o outlier & 57.32   & 43,77 & 75.10  & 40.17 & 54.09  \\ \hline
		\end{tabular}
	\end{adjustbox}
	
\end{table}


However, monitoring outlier patterns incurs non-negligible overhead due to full-tensor traversal and system queries. For example, when tracking outliers across large neural network layers, the process may require scanning through millions of tensor elements and frequently querying the system for updates. This results in significant computational and memory costs. Consequently, there is a trade-off: frequent tracking improves policy accuracy by capturing real-time changes in the model’s behavior, but at the expense of increasing the overall computational burden and slowing down training. On the other hand, infrequent tracking reduces the overhead and allows for faster processing, but it comes at the risk of missing critical shifts in the model's performance, leading to potentially suboptimal decisions. For instance, in large language models, this could mean missing key activations that impact model convergence or accuracy. \textit{Thus, determining an appropriate tracking frequency is key to balancing accuracy and efficiency, ensuring that the system remains responsive while avoiding excessive computational costs.}

\textbf{Determining the tracking frequency.} We have a new observation that the number of outliers fluctuates significantly during the early training period but stabilizes in the later stages, as shown in Figure~\ref{fig:outlier}.
\textit{Thus, we design to set a high tracking frequency in the early training stage and lower it in the later stages.} 
In \pname{}, we apply the Exponential Backoff Algorithm to determine the tracking interval (e.g., $1, 2,...2^n$ iteration), minimizing tracking overhead.
Thus, we label the training iterations as either normal iterations  or tracking iterations.

\begin{figure*}
	\centering
	\includegraphics[width=7in]{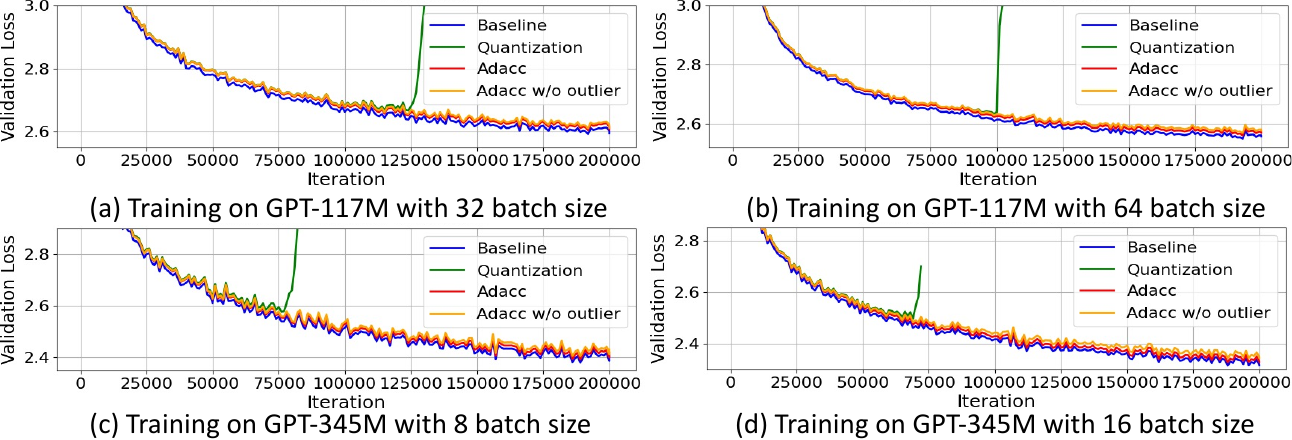}
	\caption{Training loss curve on GPT-117M and GPT-345M.}
	\label{fig:loss}
\end{figure*}

\textbf{Policy evolution mechanism.} 
During training, at each tracking iteration, the framework re-evaluates outliers in the Linear, LayerNorm, and GELU layers and inputs the updated tensor information (e.g., $CRate_i$) into the MILP cost model. Additionally, the framework allows users to customize which layers they want to re-evaluate, providing flexibility in targeting specific layers for dynamic adjustments during the training process. If the memory optimization policy changes, the framework adjusts it and continues training with the updated policy.

\section{Evaluation}
\label{sec:eval}

\subsection{Evaluation Setup}
\label{sec:setup}

\textbf{Platform.} We evaluate \pname{} on a V100  node,  which  is equipped with 256GB DRAM, 2 Intel Xeon Gold 6130 CPUs, and 8 32GB Tesla V100 GPUs interconnected via NVLink. 

\textbf{Baseline.} We compare Lynx with Megatron-LM~\cite{megatron-ckpt}, Quantization, and Baseline~\cite{megatron}. (1) \textit{Megatron-LM}, a widely used framework for large model training, supports full recomputation, checkpointing only transformer layer inputs and discarding other activations.
(2)  \textit{Quantization} compresses data directly from FP16 to INT4.
(3) \textit{Baseline} refers to training without any memory optimization, maintaining full model accuracy.
(4) \textit{Adacc w/o outlier } is Adacc with no outlier extracttion.
We apply the ZeRO technique to train all the models~\cite{Zero-SC20}.

\textbf{Workloads.} We train several GPT-like~\cite{GPU2-yu} models based on the Transformer architecture, on the pile dataset~\cite{the-pile}.

\subsection{General Result}
\label{sec:throughput}

\begin{figure}[t]
	\centering
	\subfigure[GPT-117M.]{
		\includegraphics[width=2.4in,height=1.2in]{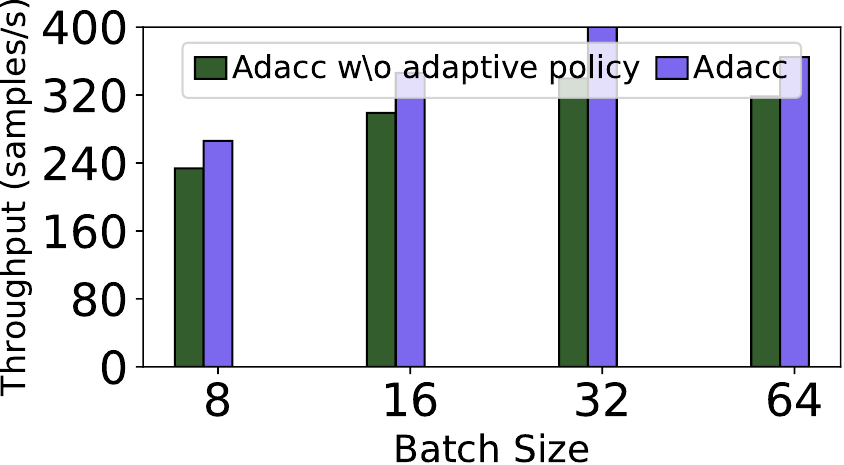}
		\label{fig:dy1}
	}
	\vspace{-0.2in}
	\subfigure[GPT-345M.]{
		\includegraphics[width=2.4in,height=1.2in]{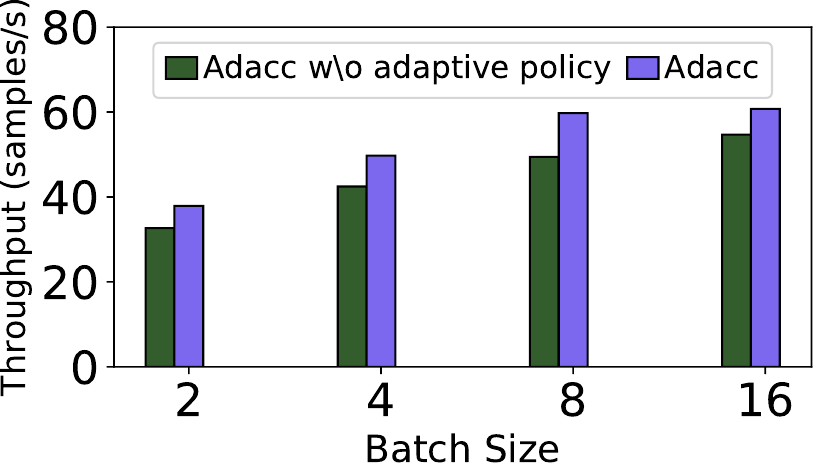}
		\label{fig:dy2}  }
	\caption{The effectiveness of the adaptive policy evolution.}
	\label{fig:dynamic}
	\vspace{-0.2in}
\end{figure}

\textbf{Training throughput.} 
To evaluate the effectiveness of \pname{}, we measure the DNN training throughput across different memory optimization approaches on two small GPT models and one large GPT model, as shown in Figure~\ref{fig:throughput}.
There are several observations. (1) 
\pname{} has the best training performance among counterparts. Specifically, \pname{} improves the throughput by 1.01$\times$-1.37$\times$ and 1.09$\times$-1.28$\times$
compared to Full Recomputation and Quantization.
(2) \pname{} adaptively selects the optimal strategy based on memory requirements to maximize training throughput. For example, with a small batch size, GPU memory is sufficient, and \pname{} introduces no additional overhead.
In contrast, Full Recomputation and Quantization only provide open and closed options, potentially leading to excessive memory optimization.
(3) For GPT-345M, as the batch size increases (e.g., from 1 to 15), the throughput of \pname{} initially rises  and then begins to decrease.
The performance gains from increased GPU utilization with larger batch training outweigh the overhead introduced by compression and recomputation.
(4) 
As the batch size increases, leading to higher GPU memory demands, \pname{} tends to recompute more tensors (the policy approaches Full Recomputation) while consistently maintaining higher throughput than Full Recomputation.

\textbf{Memory footprint reduction.}
We use the batch size to represent the degree of memory footprint reduction.
Table~\ref{tbl:maxbatch} presents the maximum batch size for different counterparts.
We observe that \pname{} always achieves the maximum batch size. 
Compared to Baseline and
Quantization, \pname{} promotes the maximum batch size by up to 2.38$\times$-7.62$\times$ and 1.38$\times$-3.2$\times$. This is because Baseline trains models without optimization techniques, while Quantization incurs significant space overhead by storing compressed data for all layers.
%



\subsection{Accuracy}
\label{sec:accuracy}
We explore the impact of \pname{}'s design on model accuracy, presenting both loss values and downstream task accuracy.

\textbf{Evaluations on validation loss.}
We show the loss curve of validation dataset on GPT-117M and GPT-345M across different batch sizes.
Figure~\ref{fig:loss} shows that \pname{} achieves comparable loss to the baseline, with only a 0.46\%-0.5\% difference in loss across two models.
Moreover, Quantization performs well in the early stages of training but fails to converge in the later stages due to significant data accuracy loss.

\textbf{Evaluations on downstream tasks.}
We evaluate zero-shot downstream tasks of our pretrained GPT-117M and 345M models.
 Each models is trained using two different batch sizes respectively,
  and applied different memory optimization methods to observe the accuracy of our pre-trained models on zero-shot tasks.
   As shown in Table~\ref{tbl:downstream}, our method has less than 0.5\% accuracy degradation compared to the Baseline (oracular method),
    while Quantization has average 39\% accuracy loss.

\subsection{Ablation Study}
\label{sec:ablation}

\textbf{The effectiveness on layer-specific compression.}
We investigate the efficiency of layer-specific compression in \pname{}.
 As shown in Figure~\ref{fig:loss}, our layer-specific compression achieves better convergence efficiency compared to \pname{} without outlier compression.
  Specifically, layer-specific compression reduces the loss by 0.2\% and 0.31\% on GPT-117M for batch sizes of 32 and 64, respectively. For the larger GPT-345M,
   \pname{} reduces the loss by 0.48\% and 0.7\% for batch sizes of 8 and 16, respectively.

\textbf{The effectiveness on adaptive policy evolution.}
We evaluate \pname{} with and without adaptive policy evolution mechanism.
We use two models with 117M and 345M parameters across four batch sizes.
Figure~\ref{fig:dynamic} illustrates that the throughput with the \pname{} adaptive policy is increased by 1.13$\times$-1.18$\times$ and 1.11$\times$-1.2$\times$ on GPT-117M and GPT-345M, respectively.

\section{Conclusion}

In this paper, we propose the \pname{} framework for large language model training with data compression and activation checkpointing memory optimization techniques.
First, we design layer-specific compression algorithms that account for outliers in LLM tensors, ensuring high training accuracy. Second, we propose an MILP-based scheduling policy to optimize memory for each tensor. Third, we introduce an adaptive policy evolution mechanism that adjusts policies during training to improve throughput.
The results show that \pname{} outperforms the SOTA method by up to 1.37$\times$.

	\bibliography{refs}
	
	\bibliographystyle{ACM-Reference-Format}

	\appendix
	
\end{document}